\documentclass{article}
\usepackage{spconf,amsmath,amssymb,graphicx,booktabs,xspace}
\usepackage[table]{xcolor}
\usepackage{enumitem}
\usepackage[hidelinks]{hyperref}

\definecolor{best}{RGB}{210, 235, 255}
\definecolor{second}{RGB}{255, 240, 210}
\newcommand{\best}[1]{\textbf{#1}}
\newcommand{\second}[1]{\underline{#1}}
\newcommand{\sysName}{COMPASS\xspace}

\title{COMPASS: Fusion-Matched Supervision for Missing-Modality Human Sensing}

\name{\begin{tabular}{c}
      Hao Wang$^{1,2*}$, Yanyu Qian$^{3,2*}$, Pengcheng Weng$^{1,2*}$, Zixuan Xia$^{1,2}$,\\
      William Dan$^{2}$, Yangxin Xu$^{2}$, Fei Wang$^{2\dagger}$\end{tabular}%
      \thanks{$^{*}$Equal contribution. $^{\dagger}$Corresponding author: feynmanw@xjtu.edu.cn}}
\address{$^{1}$Universit\"at Bern \quad $^{2}$Xi'an Jiaotong University \quad $^{3}$Nanyang Technological University}

\begin{document}
\ninept
\maketitle
\suppressfloats[t]

\begin{abstract}
Multimodal human activity recognition (HAR) and human pose estimation (HPE) must cope with modalities missing at inference. Completions generated for a fixed fusion model must preserve the fusion readout: what fusion uses from each modality. We propose \sysName, a completion-and-fusion framework in which each modality occupies a fixed slot filled by an observed representation or a completion inferred from available inputs. Its core principle, fusion-matched supervision, supervises completions with same-sample real targets at readout granularity: token means for HAR, where fusion averages tokens, and per-joint representations for HPE, where fusion preserves joint structure. Trained with complete multimodal samples, \sysName improves subset-averaged performance over the strongest baselines on XRF55 and MM-Fi for HAR and on MM-Fi for HPE. At fixed architecture, fusion-matched supervision outperforms full-token and centered-token matching in HAR, and joint-mean matching and no matching in HPE. The code is available at: {\urlstyle{same}\url{https://github.com/haowangcoder/COMPASS}}.
\end{abstract}


\section{Introduction}
\label{sec:intro}

\begin{figure}[t]
\centering
\includegraphics[width=0.85\columnwidth]{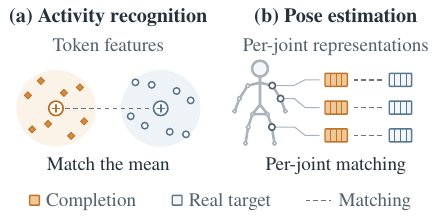}
\caption{Fusion-matched supervision in \sysName: \textbf{(a)}~HAR matches token means (circled crosses); \textbf{(b)}~HPE matches per-joint representations, indexed by the skeleton. Dashed links match completions to same-sample real targets of the missing modality, used only during training. Positions and counts are schematic.}
\label{fig:readout}
\end{figure}

Sensor failure, occlusion, hardware limitations, unstable communication, or privacy restrictions can leave modalities unavailable at inference for human activity recognition (HAR) and human pose estimation (HPE)~\cite{Wu2024MissingModalitySurvey}. Missing-modality methods adapt to available inputs or recover unavailable representations. Fusion-side approaches include subset-robust architectures~\cite{Ma2022RobustMissing}, adaptive fusion over varying sensor subsets~\cite{Chen2025XFi}, missingness-aware prompts~\cite{lee2023multimodal,jang2024towards}, and teacher distillation for missing-modality robustness~\cite{weng2026purify}; more broadly, Xia et al.~\cite{xia2026diverse} apply geometric constraints to regularize multimodal embedding spaces. Completion-side methods reconstruct missing features~\cite{Tran2017CRA,Zhao2021MMIN}, use cross-modal translation~\cite{Pham2019MCTN}, or model shared and imputed representations~\cite{Wang2023ShaSpec,Ma2021SMIL}.

Representation completion generates missing features for a fixed fusion model~\cite{Ma2021SMIL,Reza2025CMPT}, and methods differ in how these features are supervised. CMPT~\cite{Reza2025CMPT} aligns a proxy, computed from the available modality's tokens, with the missing modality's class token and adds it in fusion. ActionMAE~\cite{Woo2023ActionMAE} reconstructs a missing modality token and passes it alongside observed tokens into a fusion unit. Kim and Kim~\cite{Kim2024MissingModalityPrediction} predict missing embeddings via prompt tuning and train the predictor with VICReg~\cite{Bardes2022VICReg}. CMPT and ActionMAE supervise one token per modality, which is also what their fusion uses. Once completions carry more structure than fusion reads, or fusion keeps per-joint structure, the supervision target becomes a design choice: what structure should completions preserve, and how should supervision relate to what fusion reads?

The answer depends on the \emph{fusion readout}, the part of each modality's features that fusion uses. Our HAR fusion mean-pools a modality's tokens, so token sets with the same mean give the same readout (Fig.~\ref{fig:readout}(a)); full-token matching also constrains differences that the readout discards. Our HPE fusion keeps one vector per joint, so averaging over joints would discard structure that pose regression needs (Fig.~\ref{fig:readout}(b)). Completions should therefore be supervised at the readout's granularity, as in CMPT's alignment to the class token used in fusion.

We propose \sysName, a completion-and-fusion framework built around \emph{fusion-matched supervision}. Each modality occupies a fixed slot filled by its observed representation or a completion inferred from available inputs; training uses complete multimodal samples, whose hidden modalities supply same-sample real targets. Experiments on XRF55~\cite{Wang2024XRF55} and MM-Fi~\cite{Chen2023MMFi} show that \sysName achieves 84.0\% and 73.8\% subset-averaged HAR accuracy and 85.3\,mm mean per-joint position error (MPJPE) for MM-Fi HPE, improving on the strongest baselines by 3.6 points, 5.3 points, and 3.7\,mm, respectively.

\begin{figure*}[t]
\centering
\includegraphics[width=0.82\textwidth]{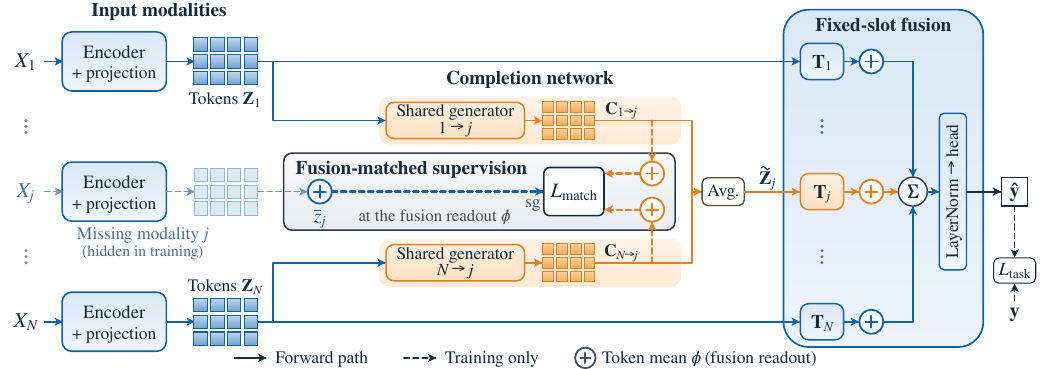}
\caption{Fusion-matched supervision in \sysName (HAR). In training, modality $j$ is hidden but still encoded, and its real features never fill its own slot; their mean $\bar z_j=\phi(\mathbf{Z}_j)$ supervises each source's completion mean before averaging, through the same token-mean readout $\phi$ (circled crosses) that fusion applies to every slot. The completions $\mathbf{C}_{i\to j}$ are averaged into $\tilde{\mathbf{Z}}_j$, which fills slot $j$. Dashed paths run only in training, and $\mathrm{sg}$ stops target gradients. VICReg on real representations is omitted; token counts are schematic.}
\label{fig:architecture}
\end{figure*}

The main contributions of this work are as follows:
\begin{itemize}[leftmargin=*,topsep=2pt,itemsep=1pt,parsep=0pt]
\item We introduce fusion-matched supervision for missing-modality human sensing, matching each completion to its same-sample real target at the granularity the fusion readout retains.
\item We design two completion networks: a shared generator conditioned on source and target modality for HAR, and a joint-structured completion network for HPE.
\item Experiments demonstrate that \sysName improves subset-averaged performance over the strongest baselines on both tasks; with the completion architecture held fixed, fusion-matched supervision outperforms full-token and centered-token matching in HAR, and both coarser joint-mean matching and no matching in HPE (Sec.~\ref{sec:analysis}).
\end{itemize}

\section{Method}
\label{sec:method}

\subsection{Completion and Fusion}
\label{sec:setup}
\label{sec:completion}

Given $N$ modalities indexed by $\mathcal{V}=\{1,\dots,N\}$, let $X_m$ denote the input of modality $m$ and $\mathbf{y}$ the ground truth, an activity label for HAR or 3D joint positions for HPE. Training samples are complete; at test time, any nonempty subset $\mathcal{O}\subseteq\mathcal{V}$ may be observed, with $\mathcal{M}=\mathcal{V}\setminus\mathcal{O}$ missing. The goal is one model per task that predicts $\mathbf{y}$ from $\{X_m\}_{m\in\mathcal{O}}$ for every nonempty $\mathcal{O}$.

Each modality occupies a fixed fusion \emph{slot}, containing its observed representation $\mathbf{Z}_m$ or a completion $\tilde{\mathbf{Z}}_m$ generated from $\mathcal{O}$:
\begin{equation}
\mathbf{T}_m =
\begin{cases}
\mathbf{Z}_m, & m \in \mathcal{O}, \\
\tilde{\mathbf{Z}}_m, & m \in \mathcal{M}.
\end{cases}
\label{eq:slot}
\end{equation}
The \emph{fusion readout} $\Phi$ maps each slot to its fusion contribution. We sum these readouts, normalize across feature dimensions (LayerNorm), and apply a task head $h$:
\begin{equation}
\mathbf{f}=\sum_{m=1}^{N}\Phi(\mathbf{T}_m),\qquad
\hat{\mathbf{y}}=h\bigl(\mathrm{LayerNorm}(\mathbf{f})\bigr).
\label{eq:fusion}
\end{equation}

\noindent\textbf{HAR (Fig.~\ref{fig:architecture}).}
Modality-specific encoders and projections map each $X_m$ to $K$ tokens $\mathbf{Z}_m\in\mathbb{R}^{K\times d}$. The readout is the token mean, $\Phi(\mathbf{A})=\phi(\mathbf{A})=\frac{1}{K}\sum_k\mathbf{A}_k$, and $h$ is a linear classifier. A shared generator serves all source--target pairs: a single-layer, four-head Transformer~\cite{Vaswani2017Transformer} takes source tokens tagged with a learned source-modality embedding, concatenated with $K$ learned queries specific to the target modality. Outputs at the query positions form the completion $\mathbf{C}_{i\to j}\in\mathbb{R}^{K\times d}$, in the same form as an observed slot. Each observed modality $i$ generates a completion for missing modality $j$; their average fills the slot:
\begin{equation}
\tilde{\mathbf{Z}}_j = \frac{1}{|\mathcal{O}|}\sum_{i\in\mathcal{O}} \mathbf{C}_{i\to j}.
\label{eq:completion}
\end{equation}
Parameters grow linearly in $N$, and completing $\mathcal{M}$ from $\mathcal{O}$ takes $|\mathcal{O}|\,|\mathcal{M}|$ generator passes.

\noindent\textbf{HPE.}
Encoder features are projected to dimension $d$. The readout $\Phi(\mathbf{A})=\mathbf{A}$ preserves each joint vector, and $h$, a shared multilayer perceptron (MLP), predicts its 3D coordinates. A shared cross-attention layer uses 17 learned joint queries, each augmented with a modality embedding, to convert a modality's tokens into one vector per joint. This \emph{joint binding} supplies observed slots and real targets for matching.

For a missing modality, a joint-structured completion network lets the joint queries attend to all observed source tokens together; each of its two layers applies cross-attention to these tokens, self-attention restricted to each joint and its skeleton neighbors, and a feed-forward network. The resulting $17\times d$ representation fills the missing slot. Fusion sums the five modality slots joint by joint.

\subsection{Fusion-Matched Supervision}
\label{sec:objectives}
A modality $j$ hidden from a complete training sample still has real features $\mathbf{Z}_j$, which serve as the target. For completion $\mathbf{S}$ and same-sample real target $\mathbf{H}$, the matching distance is
\begin{equation}
d_{\Phi}(\mathbf{S},\mathbf{H})=
\frac{1}{r_{\Phi}}\bigl\|\Phi(\mathbf{S})-\mathrm{sg}[\Phi(\mathbf{H})]\bigr\|_F^2.
\label{eq:readout_distance}
\end{equation}
Here $\mathrm{sg}$ stops target gradients, $\|\cdot\|_F^2$ sums squared entries, and $r_{\Phi}$ counts readout rows: one for HAR and 17 for HPE.

\noindent\textbf{HAR: matching token means.}
Let $\mathcal{D}(\mathbf{A})=\mathbf{A}-\mathbf{1}\phi(\mathbf{A})^{\top}$ denote the centered tokens. The full-token squared error decomposes as
\begin{equation}
\begin{aligned}
\frac{1}{K}\|\mathbf{S}-\mathbf{H}\|_F^2
&=\|\phi(\mathbf{S})-\phi(\mathbf{H})\|_2^2\\
&\quad+\frac{1}{K}\|\mathcal{D}(\mathbf{S})-\mathcal{D}(\mathbf{H})\|_F^2.
\end{aligned}
\label{eq:identity}
\end{equation}
Centered tokens sum to zero, eliminating the cross term. The first term measures fusion-readout error; the second constrains differences in the readout's \emph{null space}. With other slots fixed, these zero-mean differences leave Eq.~\eqref{eq:fusion} unchanged. Since $\Phi=\phi$, $d_{\Phi}$ matches only the first term; Sec.~\ref{sec:analysis} compares this choice with alternatives.

We supervise each source-specific completion before source averaging:
\begin{equation}
\mathcal{L}_{\mathrm{match}}=\sum_{j\in\mathcal{M}}\frac{1}{|\mathcal{O}|}
\sum_{i\in\mathcal{O}}d_{\Phi}(\mathbf{C}_{i\to j},\mathbf{Z}_j).
\label{eq:matching}
\end{equation}
For each missing modality, the average per-source loss equals the loss of the averaged completion plus the mean squared deviation of source token means from their average. Per-source supervision thus also penalizes sources that disagree even when their errors cancel on average, which matters because at test time a single source may be all that is observed.

\noindent\textbf{HPE: per-joint matching.}
Completion jointly uses all observed sources, contributing one distance $d_{\Phi}(\tilde{\mathbf{Z}}_j,\mathbf{Z}_j)$ per missing modality. Since $\Phi$ is the identity, $d_{\Phi}$ compares all 17 joint vectors. Joint binding of hidden real tokens supplies $\mathbf{Z}_j$. To stabilize matching, we divide both representations by the target's root-mean-square (RMS) vector norm.

\subsection{Training Objectives}
\label{sec:training}
We hide random subsets of modalities from complete training samples to simulate missing sensors. Hidden modalities never fill their own slots and supply the matching targets. Each training step therefore encodes all $N$ modalities, whereas inference encodes only observed ones.

HAR also applies VICReg~\cite{Bardes2022VICReg} to the means of all real modality representations, including those hidden from prediction. Its variance, covariance, and invariance terms keep variation across samples, decorrelate dimensions, and pull same-sample modalities together. Averaging over modality pairs gives $\mathcal{L}_{\mathrm{space}}=\mu_{\mathrm{inv}}\ell_{\mathrm{inv}}+\mu_{\mathrm{var}}\ell_{\mathrm{var}}+\mu_{\mathrm{cov}}\ell_{\mathrm{cov}}$. This regularizer acts on real representations, whereas $\mathcal{L}_{\mathrm{match}}$ supervises generated completions against stop-gradient targets. Kim and Kim~\cite{Kim2024MissingModalityPrediction} instead train their embedding predictor with VICReg.

The total objective is
\begin{equation}
\mathcal{L} = \mathcal{L}_{\mathrm{task}} + \lambda_c \mathcal{L}_{\mathrm{match}} + \lambda_s \mathcal{L}_{\mathrm{space}}.
\label{eq:total}
\end{equation}
For HAR, $\mathcal{L}_{\mathrm{task}}$ is cross-entropy with label smoothing~\cite{Szegedy2016LabelSmoothing}; for HPE, it is MPJPE, the average Euclidean distance between predicted and true 3D joint positions. The weights $\lambda_c$ and $\lambda_s$ control completion matching and real-representation regularization, respectively.

\section{Experiments}
\label{sec:experiments}

\begin{table}[t]
\centering
\caption{XRF55 HAR accuracy~(\%). $^\dagger$: published~\cite{Chen2025XFi}; other results are three-seed means. Avg.: mean over all seven subsets; Std.: its standard deviation across seeds. Best/second-best are bold/underlined.}
\label{tab:xrf55}
\setlength{\tabcolsep}{3pt}
\renewcommand{\arraystretch}{0.75}
\begin{tabular}{@{}lcccccc@{}}
\toprule
Observed & Feat.$^\dagger$ & X-Fi$^\dagger$ & SMIL & PTA & CMPT & \sysName \\ \midrule
R & 82.1 & 83.9 & 88.2 & 86.1 & \second{88.8} & \best{90.7} \\
W & 77.8 & 55.7 & 77.4 & 61.9 & \second{78.6} & \best{83.3} \\
RF & 42.2 & 42.5 & 44.3 & 42.9 & \second{44.9} & \best{48.3} \\
R+W & 86.8 & 88.2 & \second{91.3} & 87.6 & 89.7 & \best{94.9} \\
R+RF & 71.4 & 86.5 & 89.2 & 88.0 & \second{89.7} & \best{91.5} \\
W+RF & 55.6 & 58.1 & 72.8 & 58.3 & \second{81.7} & \best{84.4} \\
R+W+RF & 70.6 & 89.8 & \second{92.1} & 89.4 & 89.4 & \best{95.1} \\
\midrule
Avg. & 69.5 & 72.1 & 79.3 & 73.5 & \second{80.4} & \best{84.0} \\
Std. & -- & -- & 0.5 & 1.6 & 0.4 & 0.2 \\
\bottomrule
\end{tabular}
\end{table}

\begin{table}[t]
\centering
\caption{MM-Fi HAR accuracy~(\%); notation as in Table~\ref{tab:xrf55}, with Avg.\ over all 15 subsets.}
\label{tab:mmfi}
\setlength{\tabcolsep}{3pt}
\renewcommand{\arraystretch}{0.75}
\begin{tabular}{@{}lcccccc@{}}
\toprule
Observed & Feat.$^\dagger$ & X-Fi$^\dagger$ & SMIL & PTA & CMPT & \sysName \\ \midrule
I & 25.3 & 26.5 & 27.4 & 28.0 & \second{28.1} & \best{43.2} \\
D & \second{49.9} & 48.1 & 48.1 & 48.1 & 48.2 & \best{54.4} \\
L & 63.9 & 52.7 & 41.3 & \second{72.9} & 20.4 & \best{80.0} \\
R & \second{85.0} & \best{85.7} & 81.6 & 82.9 & 83.2 & 83.2 \\
I+D & 45.2 & \second{45.3} & 44.3 & 43.9 & 40.9 & \best{53.5} \\
I+L & 23.8 & 35.2 & 38.7 & \second{51.5} & 28.1 & \best{70.1} \\
I+R & 66.9 & 73.4 & 73.0 & 80.2 & \best{83.5} & \second{80.9} \\
D+L & 49.7 & 51.6 & 53.8 & \second{62.1} & 48.1 & \best{72.2} \\
D+R & 74.1 & 79.8 & 79.1 & \second{83.9} & \best{85.2} & 83.6 \\
L+R & 85.6 & \second{88.7} & 84.8 & 88.6 & 83.2 & \best{89.6} \\
I+D+L & 43.0 & 48.7 & 49.8 & \second{55.2} & 41.8 & \best{67.0} \\
I+D+R & 69.1 & 70.7 & 70.3 & \best{78.2} & 52.4 & \second{76.4} \\
I+L+R & 68.2 & 77.8 & 77.7 & \second{84.8} & 83.2 & \best{85.6} \\
D+L+R & 72.9 & 80.5 & 80.8 & \best{86.4} & 84.9 & \second{86.0} \\
I+D+L+R & 72.6 & 72.2 & 73.5 & \best{81.3} & 49.2 & \second{81.1} \\
\midrule
Avg. & 59.7 & 62.5 & 61.6 & \second{68.5} & 57.4 & \best{73.8} \\
Std. & -- & -- & 4.8 & 1.4 & 0.3 & 0.3 \\
\bottomrule
\end{tabular}
\end{table}

\subsection{Experimental Setup}
\label{sec:exp_setup}

\noindent\textbf{Datasets and protocols.}
We evaluate \sysName on two public multimodal human sensing datasets: XRF55 for HAR and MM-Fi for HAR and HPE. XRF55~\cite{Wang2024XRF55} contains 42.9K radio-frequency samples from 39 subjects for 55 activities. We use its original split and all seven nonempty subsets of mmWave heatmaps~(R), WiFi channel state information (WiFi-CSI)~(W), and RFID phase series~(RF). MM-Fi~\cite{Chen2023MMFi} contains over 320K synchronized frames from 40 subjects. Following X-Fi~\cite{Chen2025XFi}, both tasks use MM-Fi's S1 random split of video samples (80/20 per activity, split seed 0). HAR covers 27 activities and all 15 subsets of RGB~(I), depth~(D), LiDAR~(L), and mmWave~(R). HPE adds WiFi-CSI~(W), giving five modalities, 17 joints, and 31 subsets. We report HAR top-1 accuracy and, for HPE, MPJPE and Procrustes-aligned MPJPE (PA-MPJPE) in mm. Subset-averaged results weight all nonempty subsets equally.

\noindent\textbf{Baselines.}
\looseness=-1 For HAR, feature-level fusion (Feat.) and X-Fi~\cite{Chen2025XFi} are published under X-Fi's protocol; SMIL~\cite{Ma2021SMIL}, PTA~\cite{weng2026purify}, and CMPT~\cite{Reza2025CMPT} are adapted with trainable encoders: SMIL imputes missing features, PTA applies meta-weighted distillation, and CMPT regresses one proxy per missing modality onto that modality's real token mean. For HPE, decision-level fusion (Decision), feature-level fusion (Feature), and X-Fi are published S1 results; SMIL, PTA, and CMPT are adapted to our frozen-feature pipeline, with SMIL fusing observed features and CMPT as a task-trained global completion.

\noindent\textbf{Implementation details.}
Encoders follow X-Fi, initialized from pretrained per-modality models: ResNet-18~\cite{He2016ResNet} for RGB, depth, and XRF55 radio-frequency inputs, Point Transformer~\cite{Zhao2021PointTransformer} for point clouds, and X-Fi's WiFi-CSI encoder for HPE. Our main and target-comparison runs use AdamW~\cite{Loshchilov2019AdamW} with base learning rate $10^{-4}$ for all trainable parameters and report means and standard deviations over seeds 0, 1, 2. HAR uses $K=32$ and $d=512$ and fine-tunes encoders for 100 epochs (batch 32) on XRF55 and 50 epochs (batch 16, capped at 1000 batches/epoch) on MM-Fi; weight decay 0.05 and $5\times10^{-4}$ respectively, five-epoch linear warmup, polynomial decay, final-epoch checkpoints, $\lambda_c=0.016$, $\lambda_s=0.3$, and $(\mu_{\mathrm{inv}},\mu_{\mathrm{var}},\mu_{\mathrm{cov}})=(0,25,1)$ on XRF55 and $(5,25,1)$ on MM-Fi. With probability $p_{\mathrm{drop}}=0.7$, each HAR training sample keeps 1 to $N-1$ modalities, drawing the count and then the subset uniformly; otherwise all modalities are observed. HPE samples one of the 31 nonempty subsets uniformly per batch. HPE uses $d=256$ and freezes encoders, training 60 epochs of 1000 iterations (batch 64, weight decay 0.01, $\lambda_c=0.1$); the target RMS norm is computed per modality over the batch and joints without gradient, and the checkpoint minimizes MPJPE on a 10\% training holdout.

\subsection{Main Results}
\label{sec:exp_main}

\begin{table}[t]
\centering
\setlength{\tabcolsep}{2.3pt}
\renewcommand{\arraystretch}{0.75}
\caption{MM-Fi HPE MPJPE (mm, $\downarrow$) by observed modality count; Avg.: all 31 subsets, $\pm$ std.\ over three seeds; $^\dagger$: published~\cite{Chen2025XFi}. Best/second-best among main methods are bold/underlined (last two rows: \sysName ablations).}
\label{tab:hpe}
\begin{tabular}{@{}lcccccc@{}}
\toprule
Method / $|\mathcal{O}|$ & 1 & 2 & 3 & 4 & 5 & Avg. \\ \midrule
Decision$^\dagger$ & 150.7 & 124.1 & 113.0 & 106.8 & 103.0 & 121.3 \\
Feature$^\dagger$ & 150.7 & 139.2 & 131.5 & 139.0 & 128.9 & 138.2 \\
X-Fi$^\dagger$ & 143.2 & 105.0 & 92.6 & 87.9 & 83.7 & 103.7 \\
SMIL & \second{130.2} & 92.2 & 79.9 & 74.3 & 72.2 & 90.8$\pm$1.5 \\
PTA & 131.1 & \second{90.4} & \second{77.2} & \second{71.5} & \best{69.2} & \second{89.0}$\pm$1.1 \\
CMPT & 131.1 & 90.9 & 77.4 & 71.8 & \second{71.9} & 89.3$\pm$2.9 \\
\midrule
\sysName & \best{122.1} & \best{86.4} & \best{74.7} & \best{70.5} & \best{69.2} & \best{85.3}$\pm$0.3 \\
\midrule
\enspace No matching & 130.1 & 92.2 & 79.0 & 75.8 & 78.8 & 91.0$\pm$1.4 \\
\enspace Joint-mean matching & 127.5 & 89.7 & 76.8 & 71.8 & 69.8 & 88.1$\pm$0.9 \\
\bottomrule
\end{tabular}

\par\vspace{6pt}
\caption{MM-Fi HPE PA-MPJPE (mm, $\downarrow$); notation as in Table~\ref{tab:hpe}.}
\label{tab:hpe_pa}
\begin{tabular}{@{}lcccccc@{}}
\toprule
Method / $|\mathcal{O}|$ & 1 & 2 & 3 & 4 & 5 & Avg. \\ \midrule
Decision$^\dagger$ & 80.9 & 70.0 & 65.8 & 63.7 & 62.4 & 69.2 \\
Feature$^\dagger$ & 80.9 & 73.8 & 71.3 & 71.9 & 72.5 & 73.8 \\
X-Fi$^\dagger$ & 77.4 & 59.7 & 52.3 & 49.6 & 47.6 & 58.2 \\
SMIL & \second{76.5} & 60.5 & 53.7 & 50.8 & 49.7 & 59.0$\pm$0.6 \\
PTA & 80.4 & 60.4 & 52.5 & 49.1 & 47.6 & 58.9$\pm$0.7 \\
CMPT & 77.7 & \second{59.5} & \second{51.4} & \second{48.1} & \best{47.3} & \second{57.6}$\pm$1.1 \\
\midrule
\sysName & \best{73.1} & \best{56.6} & \best{49.9} & \best{48.0} & \second{47.4} & \best{55.4}$\pm$0.2 \\
\midrule
\enspace No matching & 76.2 & 58.6 & 50.6 & 48.4 & 52.4 & 57.0$\pm$0.2 \\
\enspace Joint-mean matching & 76.0 & 58.8 & 51.3 & 49.0 & 48.5 & 57.3$\pm$0.4 \\
\bottomrule
\end{tabular}

\end{table}

On XRF55 (Table~\ref{tab:xrf55}), \sysName achieves 84.0$\pm$0.2\% subset-averaged accuracy, improving on CMPT (80.4\%) by 3.6 points, and ranks first on all seven subsets. On MM-Fi HAR (Table~\ref{tab:mmfi}), it achieves 73.8$\pm$0.3\%, improving on PTA (68.5\%) by 5.3 points; it ranks first on all seven subsets without mmWave and stays within 2.6 points of the leader on the eight with mmWave. On MM-Fi HPE (Tables~\ref{tab:hpe} and~\ref{tab:hpe_pa}), it achieves 85.3$\pm$0.3\,mm MPJPE and 55.4$\pm$0.2\,mm PA-MPJPE, improving on PTA (89.0\,mm) and CMPT (57.6\,mm) by 3.7 and 2.2\,mm, respectively.

\noindent\textbf{Efficiency.}
The shared HAR generator has 2.1M parameters, plus about 17K per modality for its source tag and target queries. One pass costs 0.28 GFLOPs; four passes on MM-Fi add 1.1 GFLOPs and 1.1\,ms to a 4.6\,ms forward pass at batch size 1 on an RTX 4090.

\subsection{Supervision and Regularization Analysis}
\label{sec:analysis}

\begin{table}[!tb]
\centering
\caption{HAR supervision and regularization ablations. Accuracy (\%) is averaged over all nonempty modality subsets; values are mean $\pm$ std.\ over three seeds. Best results are bold.}
\label{tab:supervision}
\label{tab:invariance_matching}
\begingroup
\setlength{\tabcolsep}{3pt}
\renewcommand{\arraystretch}{0.9}
\begin{tabular*}{\columnwidth}{@{\extracolsep{\fill}}llcc@{}}
\multicolumn{4}{c}{\textbf{(a) Supervision targets}} \\[2pt]
\toprule
\multicolumn{2}{@{}l}{Supervision} & XRF55 & MM-Fi \\
\multicolumn{2}{@{}l}{} & Invariance off & Invariance on \\ \midrule
\multicolumn{2}{@{}l}{No matching} & 81.63$\pm$0.34 & 60.4$\pm$2.6 \\
\multicolumn{2}{@{}l}{Full-token matching} & 82.72$\pm$0.35 & 66.9$\pm$0.3 \\
\multicolumn{2}{@{}l}{Centered-token matching} & 82.36$\pm$0.24 & 69.1$\pm$0.9 \\
\midrule
\multicolumn{2}{@{}l}{Fusion-matched (ours)} & \best{84.03}$\pm$0.23 & \best{73.8}$\pm$0.3 \\
\bottomrule
\addlinespace[6pt]
\multicolumn{4}{c}{\textbf{(b) Invariance and matching}} \\[2pt]
\toprule
Invariance & Matching & XRF55 & MM-Fi \\ \midrule
Off & Off & 81.63$\pm$0.34 & 40.28$\pm$0.16 \\
On & Off & 83.76$\pm$0.03 & 60.40$\pm$2.60 \\
Off & On & \best{84.03}$\pm$0.23 & 68.51$\pm$0.27 \\
On & On & 83.68$\pm$0.29 & \best{73.80}$\pm$0.30 \\
\bottomrule
\end{tabular*}
\endgroup
\end{table}

\noindent\textbf{Supervision targets.}
Table~\ref{tab:supervision}(a) compares supervision targets with the generator, source averaging, and regularization held fixed. Full-token matching includes both terms of Eq.~\eqref{eq:identity}; centered-token matching retains only the centered term; fusion-matched supervision retains only the token-mean term. On the six XRF55 subsets with missing modalities, fusion-matched supervision averages 82.2\%, versus 80.9\% for full-token matching and 80.4\% for centered-token matching; on MM-Fi, it leads both alternatives by at least 4.7 points. Removing matching ($\lambda_c=0$) retains the completion network but lowers accuracy on both datasets. Centered-token matching also improves over no matching, although mean pooling discards the differences it supervises.

\noindent\textbf{Invariance and matching interaction.}
Table~\ref{tab:supervision}(b) toggles VICReg's invariance term ($\mu_{\mathrm{inv}}>0$) and fusion-matched supervision ($\lambda_c>0$). Invariance aligns real representations across modalities, whereas matching aligns each completion with its same-sample real target. On MM-Fi, their combination exceeds either alone; on XRF55 it does not. The main configurations use matching on both datasets.

\noindent\textbf{HPE structure and matching.}
Per-joint matching in \sysName achieves the lowest subset-averaged MPJPE and PA-MPJPE among the HPE variants (Tables~\ref{tab:hpe} and~\ref{tab:hpe_pa}): 85.3/55.4\,mm, versus 91.0/57.0\,mm without matching and 88.1/57.3\,mm with joint-mean matching. No matching retains the completion network but removes its matching loss; joint-mean matching compares representations averaged across joints. These results support per-joint matching for the joint-preserving fusion readout.

\noindent\textbf{Loss scaling.} Fusion-matched supervision and centered-token matching use Eqs.~\eqref{eq:readout_distance} and~\eqref{eq:identity} as written with $\lambda_c=0.016$. Full-token matching uses the per-entry mean, Eq.~\eqref{eq:identity} divided by $d$, with weight 2.05 on XRF55 and 0.2 on MM-Fi. Generator depth, head count, $\lambda_s$, and $p_{\mathrm{drop}}$ are shared by the two HAR datasets.

\noindent\textbf{Weight sensitivity.} On XRF55, sweeping the full-token weight from 0.51 to 8.19 ($16\times$, seed 0) gives 81.9--82.3\%. At 8.19, its per-pair token-mean coefficient is $8.19/512\approx0.016$, approximately our $\lambda_c$; the gap to fusion-matched supervision persists. Quadrupling our MM-Fi matching weight (seed 0) leaves subset-averaged accuracy nearly unchanged (73.8 to 73.7\%) but lowers fully observed accuracy from 81.3 to 79.5\%. For HPE, $\lambda_c$ from 0.02 to 0.1 (ours: 0.1) keeps MPJPE within 84.9--85.9\,mm; it rises to 88.8 and 89.9\,mm at 0.01 and 0.3.

\section{Conclusion}
\label{sec:conclusion}
We introduced \sysName, a completion-and-fusion framework that completes missing modalities in fixed fusion slots with fusion-matched supervision. Fusion-matched supervision matches each completion to its same-sample real target at the granularity the fusion readout retains: token means for HAR and per-joint representations for HPE. Experiments show that \sysName improves subset-averaged performance over the strongest baselines on XRF55 and MM-Fi for HAR and on MM-Fi for HPE. At fixed architecture, HAR target comparisons and HPE ablations support supervising completions at the granularity retained by the fusion readout.

\clearpage
\bibliographystyle{IEEEtran}
\bibliography{main}

\end{document}